\def\year{2020}\relax
\documentclass[letterpaper]{article} 
\usepackage{aaai20}  
\usepackage{times}  
\usepackage{helvet} 
\usepackage{courier}  
\usepackage[hyphens]{url}  
\usepackage{graphicx} 
\usepackage{graphicx}  
\usepackage{tabularx, enumitem,booktabs, cellspace, multirow}
\usepackage{graphicx}
\usepackage{amsmath}
\usepackage{scalerel,amssymb}
\usepackage{color}
\usepackage{url}
\usepackage{subfigure}
\usepackage{balance}

\usepackage{algorithm}
\usepackage[noend]{algpseudocode}
\usepackage{calrsfs}

\makeatletter
\def\BState{\State\hskip-\ALG@thistlm}
\makeatother

\usepackage{xcolor}

\newcolumntype{P}[1]{>{\centering\arraybackslash}p{#1}}

\newcommand{\emu}{\mbox{\sc Emu}}
\newcommand{\emuNoLD}{\mbox{\sc Emu} w/o LD}
\newcommand{\emuNoLDCL}{\mbox{\sc Emu} w/o LD+CL}
\newcommand{\emup}{\mbox{\sc Emu-Parallel}}

\newcommand{\trustyou}{{\textit{HotelQA}}}
\newcommand{\quora}{{\textit{Quora}}}
\newcommand{\atis}{{\textit{ATIS}}}

\newcommand{\hl}[1]{#1}
\newcommand{\wctan}[1]{}
\newcommand{\wataru}[1]{}
\newcommand{\yoshi}[1]{}
\newcommand{\behzad}[1]{}
\newcommand{\todo}[1]{}

\newcommand{\superscript}[1]{\scalebox{0.75}{\ensuremath{^{\textrm{#1}}}}}
\newcommand{\sigS}{\textnormal{\superscript{*}}}
\newcommand{\sigSS}{\textnormal{\superscript{**}}}
\newcommand{\sigSSS}{\textnormal{\superscript{***}}}

\title{Emu: Enhancing Multilingual Sentence Embeddings with Semantic Specialization}

\author{Wataru Hirota\textsuperscript{\rm 1}\thanks{This work was done during an internship at
Megagon Labs.}, \Large \textbf{Yoshihiko Suhara}\textsuperscript{\rm 2}, \textbf{Behzad Golshan}\textsuperscript{\rm 2}, \textbf{Wang-Chiew Tan}\textsuperscript{\rm 2}\\
\textsuperscript{\rm 1}Osaka University, \textsuperscript{\rm 2}Megagon Labs\\
\textrm{w-hirota@ist.osaka-u.ac.jp, \{yoshi, behzad, wangchiew\}@megagon.ai}
}

\begin{document}
\maketitle


\begin{abstract}
We present \emu, a system that semantically enhances multilingual sentence embeddings.
Our framework fine-tunes pre-trained multilingual sentence embeddings using two main components: a semantic classifier and a language discriminator. The semantic classifier improves the semantic similarity of related sentences, whereas the language discriminator enhances the multilinguality of the embeddings via multilingual adversarial training.
Our experimental results based on several language pairs show that our specialized embeddings outperform the state-of-the-art multilingual sentence embedding model on the task of cross-lingual intent classification using only monolingual labeled data.
\end{abstract}

\section{Introduction}\label{sec:introduction}
Learning multilingual sentence representations~\cite{Ruder:2019:CLSurvey} is a key technique for building NLP applications with multilingual support. A primary advantage of multilingual sentence embeddings is that they enable us to train a single classifier based on a single language (e.g., English) and then apply it to other languages without using training models for those languages (e.g., German.)
Furthermore, recent advances in multilingual sentence embedding techniques~\cite{Artetxe:2018:LASER,Chidambaram:2018:CrossLingUSE} have shown to exhibit competitive performance on several downstream NLP tasks, compared to the two-stage approach that relies on machine translation followed by monolingual sentence embedding techniques.

The main challenge of multilingual sentence embeddings is that they are sensitive to textual similarity ({\it textual similarity bias}) which negatively affects the the semantic similarity of sentence embeddings~\cite{Zhu:2018:ExploringSemanticProp}.
The following example illustrates this point:
\begin{itemize}
  \setlength{\parskip}{0cm}
  \setlength{\itemsep}{0cm}
\item[] S1: \textit{What time is the pool open tonight?}
\item[] S2: \textit{What time are the stores on 5th open tonight?}
\item[] S3: \textit{When does the pool open this evening?}
\end{itemize}
\normalsize
S1 and S3 have similar intents. They ask for the opening hours of the pool in the evening.
S2 has a different intent: it asks about the opening hour of stores. We expect embeddings of sentences of the same intent to be closer (e.g., to have higher cosine similarity) to one another than embeddings of sentences with different intents. 

We tested several pre-trained (multilingual) sentence embedding models~\cite{Pagliardini:2017:sent2vec,Conneau:2017:InferSent,Artetxe:2018:LASER,Chidambaram:2018:CrossLingUSE} in both monolingual and cross-lingual settings. 
Somewhat surprisingly, every model provided lower
similarity scores between S1 and S3 (compared to S1 and S2, or S2 and S3).
This is mainly because S1 and S2 are more textually similar (because both sentences contain ``what time'' and ``tonight'') compared to S1 and S3.
This example highlights that general-purpose multilingual sentence embeddings exhibit textual similarity bias, which is 
a fundamental limitation as they may not correctly capture the semantic similarity of sentences.

Motivated by the need for sentence embeddings that better reflect the semantics of sentence, we examine {\it multilingual semantic specialization}, which tailors pre-trained multilingual sentence embeddings to handle semantic similarity.
Although prior work has developed {\it semantic specialization} methods for word embeddings~\cite{Mrksic:2017:SemanticSpecialization} and semantic and linguistic properties of sentence embeddings~\cite{Zhu:2018:ExploringSemanticProp,Conneau:2018:WhatYouCanCram}, no prior work has considered semantic specialization of multilingual sentence embeddings.

In this paper, we develop a ``lightweight'' approach for semantic specialization of multilingual embeddings that can be applied to any base model. Our approach fine-tunes a pre-trained multilingual sentence embedding model based on a classification task that considers semantic similarity.  This aligns with common techniques of pre-training methods for NLP ~\cite{Jeremy:2018:ULMFiT,Peters:2018:ELMo,Devlin:2018:BERT}.
We explore several loss functions to determine which is appropriate for the semantic specialization of cross-lingual sentence embeddings.

We found that naive choices of loss functions such as the softmax loss, which is a common choice for classification, may suffer from significant degradation of the original multilingual sentence embedding model. 

We also design \emu{} to specialize multilingual sentence embeddings using only monolingual training data as it is expensive to collect parallel training data in multiple languages. 
Our solution incorporates {\it language adversarial training} to enhance the multilinguality of sentence embeddings. 
Specifically, we implemented a language discriminator that tries to identify the language of an input sentence given its embedding and optimizes multilingual sentence embeddings to confuse the language discriminator.

We conducted experiments on three cross-lingual intent classification tasks that involves 6 languages. The results show that \emu{} successfully specializes the state-of-the-art multilingual sentence embedding techniques, namely LASER, using only monolingual training data with unlabeled data in other languages. It outperforms the original LASER model and monolingual sentence embeddings with machine translation by up to 47.7\% and 86.2\% respectively.

The contributions of the paper are as follows:
\begin{itemize}
  \setlength{\parskip}{0cm}
  \setlength{\itemsep}{0cm}
 \item We developed \emu, a system that semantically enhances pre-trained multilingual sentence embeddings\footnote{\hl{Our code is available at \url{https://github.com/megagonlabs/emu}.}}. \emu{}
incorporates multilingual adversarial training on top of fine-tuning to enhance multilinguality without using parallel sentences.
 \item We experimented with several loss functions and show that the two loss functions, namely $L_2$ constrained softmax and center loss, outperform common loss functions used for fine-tuning. 
 \item We show that \emu{} successfully specializes multilingual sentence embedding using only monolingual labeled data.
\end{itemize}

\section{Multilingual Semantic Specialization}
\begin{figure}[t]
\centering
\includegraphics[width=0.3\textwidth]{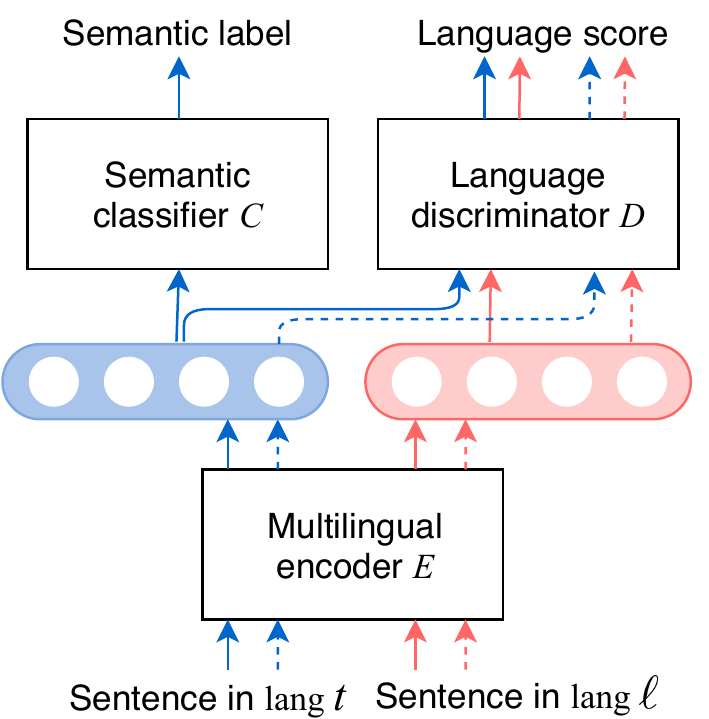}
\caption{Architecture of \emu.}
\label{fig:crosslingual_adversarial}
\end{figure}

The architecture of \emu{} is depicted in Figure~\ref{fig:crosslingual_adversarial}. There are three  main components, which we detail next: multilingual encoder $E$, semantic classifier $C$, and language discriminator $D$. The solid lines show the flow of the forward propagation for fine-tuning $C$ and $E$, and the dotted lines are that for $D$. These arrows become reversed during the backpropagation. The semantic classifier and language discriminator are only used for fine-tuning. 

After fine-tuning, \emu{} uses the fine-tuned multilingual encoder to obtain sentence embeddings for input sentences. More specifically, we expect the similarity (e.g., cosine similarity) between two related sentences in any languages to be closer to each other. We consider cosine similarity as it is the most common choice and can be calculated efficiently~\cite{Wang:2017:LearningToHash}.

\subsection{Multilingual Encoder}
A multilingual encoder is a language-agnostic sentence encoder that converts sentences in any language into embedding vectors in a common space. 
\emu{} is flexible with the choice of multilingual
encoders and their architectures. The only requirement of this component is that it encodes a sentence in any language into a sentence embedding.

In this paper, we use LASER~\cite{Artetxe:2018:LASER} as a base multilingual sentence embedding model.
LASER is a multilingual sentence embedding model that covers more than 93 languages with more than 23 different alphabets. It is an encoder-decoder model that shares the same BiLSTM encoder with max-pooling and uses \hl{byte pair encoding (BPE)}~\cite{Sennrich:2016:BPE} to accept sentences in any languages as input. The model is trained on a set of bilingual translation tasks
and is shown to have the state-of-the-art performance on cross-lingual NLP tasks including bitext mining.
We use LASER instead of multilingual models for BERT~\cite{Devlin:2018:BERT} because (1) LASER outperformed the BERT model on the XNLI task~\cite{Artetxe:2018:LASER} and (2) a LASER model can be used as a sentence encoder without any changes\footnote{A BERT model needs to be fine-tuned to use the first vector corresponding to \hl{the class symbol} [CLS] as a sentence embedding. \hl{A BERT variant for sentence embeddings also needs supervision to train a model~\cite{Reimers:2019:SentenceBERT}}.}.

\subsection{Semantic Classifier}
The semantic classifier categorizes input sentences into groups that share the same intent, such as ``seeking pool information'' or ``seeking restaurant information''.
We expect the semantic classifier to enhance multilingual sentence embeddings to better reflect the semantic similarity of related sentences, where the semantic similarity is calculated as the cosine similarity between the embeddings of the two sentences.

Additionally, we expect that learned embeddings retain semantic similarity with respect to cosine similarity. 
Thus, we propose the use of $L_2$-constrained softmax loss~\cite{Ranjan:2017:L2constrainedSL} and center loss~\cite{Wen:2016:CenterLoss}, 
which are known to be effective for image recognition tasks. To the best of our knowledge, we are the first to apply these loss functions for fine-tuning embedding models. We describe these loss functions next.

\noindent
{\bf $L_2$-constrained softmax loss} $L_2$-constrained softmax loss~\cite{Ranjan:2017:L2constrainedSL} considers hard constraints on the norm of embedding vectors on top of the softmax loss:
{
\begin{equation}
\begin{aligned}
& \text{minimize} & & - \frac{1}{M} \sum_{i=1}^M \log \frac{e^{W_{y_i}^T u_i + b_{y_i}}}{\sum_{j=1}^C e^{W_j^T u_i + b_j}}\\
& \text{subject to} & & \|u_i\|_2 = \alpha, \forall i = 1, \ldots, M,
\notag
\end{aligned}
\end{equation}
}
where $M$ denotes the number of classes, and $u_i$ and $y_i$ are $i$-th sentence embedding vector and its true label respectively.

The $L_2$ constraint ensures that embedding vectors are distributed on the hypersphere with the size of $\alpha.$ Therefore, the Euclidean distance between two vectors on the hypersphere is approximately close to its cosine distance.
This property is helpful for specializing sentence embeddings to learn semantic similarity in the form of cosine similarity.
Note that this $L_2$-constraint is different from the $L_2$ regularization term applied to the weight parameters of the output layer. In that case, the regularization term will be considered in the loss function.

To implement $L_2$-constrained softmax loss, the model additionally inserts an $L_2$-normalized layer that normalizes the encoder output $u$ (i.e., $\frac{u}{\|u\|}$) followed by a layer that scales with a hyper-parameter $\alpha$. The scaled vectors are then fed into the output layer, where the model evaluates the softmax loss.

\noindent
{\bf Center loss} The center loss~\cite{Wen:2016:CenterLoss} was originally developed for face recognition tasks to stabilize deep features learned from data. The center loss is described as follows:
{
\begin{equation}
    L_{center} = \frac{1}{2} \sum_{i=1}^m \| u_i - c_{y_i} \|_2^2,
\end{equation}
}
\noindent
where $c_{y_i}$ denotes the centroid of sentence embedding vectors of class $y_i$. The loss function forces the embedding vector of $i$-th sample toward
the centroid of the true category. 
%
Our motivation to use this loss function is to enhance the {\it intra-class compactness} of sentence embeddings. That is, we want to ensure that the sentence embeddings that have the same intent form compact clusters because other loss functions, such as the softmax loss, does not have this functionality. The center loss works as {\it cross-lingual} center loss; it enforces sentences, in any language, that belong to the same intent as a same cluster if multilingual training data are available.

We consider combining the center loss with another function with a hyper-parameter $\lambda$:
{
\begin{equation}
 L_C = L_{L_2\mbox{-}sm} + \lambda L_{center},
\end{equation}
}
\noindent
where $L_{L_2\mbox{-}sm}$ denotes the $L_2$-constrained softmax loss function.

\subsection{Language Discriminator}
The semantic classifier does not directly consider {\it multilinguality}, so the model, which is fine-tuned on a single language, may now perform worse on other languages.
To avoid this problem, we incorporate {\it multilingual adversarial learning} into the framework. Specifically, the language discriminator $D$ aims to identify the language of an input sentence given its embedding, whereas the multilingual sentence encoder $E$ incorporates an additional loss function to ``confuse'' $D$. The idea was inspired by related work that used adversarial learning for multilingual NLP models~\cite{Chen:2018:ADAN,Chen:2018:UnsupervisedMWE}. We hypothesize and our experiments show that incorporating adversarial learning also enhances the multilinguality of sentence embeddings.

The language discriminator is trained to determine whether the languages of two input embeddings are different. Simultaneously, the other part of the model is trained to confuse the discriminator. In our implementation, we use Wasserstein GAN~\cite{Arjovsky:2017:WGAN} because it is known to be more robust than the original GAN~\cite{Goodfellow:2014:GAN}.

Algorithm 1 shows a single training step of \emu. Each step consists of two training routines for language discriminator $D_t$ and the other components (multilingual sentence encoder $E$ and semantic classifier $C$). Target language $t$ denotes the language used for training (e.g., English). $t$ is randomly chosen from a training language set if multiple languages are used for training. {\it Adversarial languages} ${\mathcal{L}}$ is a set of languages that are used to retrieve adversarial sentences.
To train language discriminator $D_t$, training sentences in language $t$ and adversarial sentences from randomly chosen language $\ell \in {\mathcal{L}}$ are used to evaluate $L_{D_t}$.
Formally, the loss function for any training language $t$ is described as 
{
\begin{equation}\label{eq:adv_loss}
    L_{D_t} = L_d(1, D_t(u^t)) + L_d(0, D_t(v^\ell)),
\end{equation}}
where $L_d(\cdot, \cdot)$ is the cross entropy loss, $u_t$ and $v_\ell$ are embedding vectors (encoded by $E$) of sentences in language $t$ and language $\ell$ ($t \ne \ell$). 
Our design implements a language discriminator for each training language $t$. For instance, language discriminator $D_{t = en}$ aims to predict whether an input multilingual sentence embedding belongs to English. 

Next, labeled sentences in language $t$ and adversarial sentences $\ell$ are sampled to update the parameters of $E$ and $C$ with the fixed parameters of $D_t$. The overall loss function $L_{C+D_t}$ now takes into account the loss value of $D_t$ so that the multilingual encoder $E$ can generate multilingual sentences embeddings for sentences in languages $t$ and $\ell$, which cannot be classified by the language discriminator $D_t$. We use hyper-parameter $\gamma$ to balance the loss functions:
{
\begin{equation}\label{eq:total_loss}
    L_{C+D_t} = L_C - \gamma L_{D_t}.
\end{equation}}

\begin{algorithm}[t]
\footnotesize
\caption{Single Training Step of \emu{}}\label{alg:adversarial}
\begin{algorithmic}[1]
\Require{Training lang $t$, adversarial langs ${\mathcal{L}}$, iteration number $k$, clipping interval $c$.}
\For{$1$ to $k$}
\State Sample training sentences as $x^{t}$
\State Sample adversarial language $\ell$ from ${\mathcal{L}}$
\State Sample adversarial sentences as $x^{\ell}$
\State $u^{t} \gets E(x^{t})$; \hspace{0.5em} $v^{\ell} \gets E(x^{\ell})$
\State Evaluate loss $L_{D_t}(u^{t}, v^{\ell})$ \Comment{Eq. \ref{eq:adv_loss}}
\State Update $D_t$ parameters 
\State Clip $D_t$ parameters to $[-c, c]$
\EndFor
\State Sample training sentences and labels as $x^{t}$ and $y^{t}$
\State Sample adversarial language $\ell$ from ${\mathcal{L}}$
\State Sample adversarial sentences as $x^{\ell}$
\State $u^{i} \gets E(x^{t})$; \hspace{0.5em} $v^{\ell} \gets E(x^{\ell})$
\State Evaluate loss $L_{C+D_t}(u^{t}, v^{\ell}, y^{t})$ \Comment{Eq. \ref{eq:total_loss}}
\State Update $E$ and $C$ parameters 
\end{algorithmic}
\normalsize
\end{algorithm}

\begin{table}[t!]
\footnotesize
    \centering
    \caption{Statistics of the datasets.} 
    \begin{tabular}{c||c|c|c}\hline
                       & \trustyou{} & \atis{} & \quora{}  \\\hline
        \# of classes & 28 & 13 & 50 \\
        \# of training data & 676 & 1,195 & 1,059 \\
        \# of test data & 144 & 252 & 353 \\
        Vocab. size (en) & 977 & 626 & 1,308 \\\hline
    \end{tabular}
    \label{tab:dataset_stats}
\normalsize
\end{table}

\section{Evaluation}
We evaluated \emu{} based on the cross-lingual intent classification task.
The task is to detect the intent of an input sentence in a source language (e.g., German) based on {\it labeled sentences} associated with intent labels in a target language (e.g., English.)
We consider similarity-based intent detection, which categorizes an input sentence based on the label of the {\it nearest neighbor} sentence that has the highest cosine similarity against the input sentence.
We adopted this evaluation method since it is widely used in search-based QA systems~\cite{Pacsca:2003:OpenQA} and works robustly especially if training data are sparse.
%
An intuitive alternative for intent detection is to directly use the trained semantic classifier (see Figure~\ref{fig:crosslingual_adversarial}). We evaluated the classification results using the semantic classifier but the performance was poor. Therefore, we excluded the results from the tables.

\begin{table*}[th]
    \centering
    \caption{Experimental results (Acc@1) on three dataset. The highest performance (excluding \emup{}) is in bold and the highest performance by \emup{} is underlined. \sigS, \sigSS, and \sigSSS denote $p$-value $< 0.01$, $0.05$, and $0.10$ respectively based on the binomial proportion confidence intervals of Acc@1 values against the baseline methods.}\label{table:results_ty}
    \vspace{0.5em}
    \footnotesize                
    (a) \trustyou{}\\
    \begin{tabular}{l|l|p{0.8cm}|p{0.8cm}p{0.8cm}p{0.8cm}p{0.8cm}p{0.8cm}|p{0.8cm}p{0.8cm}p{0.8cm}p{0.8cm}p{0.8cm}}\hline
                & & & \multicolumn{5}{c|}{{\sc En} $\to$ *} & \multicolumn{5}{c}{* $\to$ {\sc En}} \\\hline
                & Method & en-en & de & es & fr & ja & zh & de & es & fr & ja & zh \\\hline
\multirow{5}{*}{\rotatebox{90}{Baseline}} & MT + sent2vec &   48.6 &   41.0 &   35.4 &   34.7 &   47.2 &   43.1 &   46.5 &   47.2 &   44.4 &   48.6 &   41.7 \\
& LASER (original)              &  55.6 &   45.1 &   48.6 &   48.6 &   47.9 &   45.1 &   43.8 &   45.8 &   50.7 &   44.4 &   49.3 \\\cline{2-13}
& Contrastive loss              &   34.0 &   19.4 &   12.5 &   22.9 &   25.7 &   21.5 &   24.3 &   18.8 &   25.7 &   24.3 &   20.1 \\
& N-pair loss                   &    27.8 &   20.8 &   22.9 &   21.5 &   20.8 &   21.5 &   24.3 &   24.3 &   25.7 &   25.0 &   20.1 \\
& Softmax loss                  &  30.6 &   13.9 &   13.9 &    7.6 &    8.3 &    7.6 &   13.2 &   24.3 &   16.0 &   20.8 &   13.9  \\\hline
\multirow{4}{*}{\rotatebox{90}{Proposed}} & \emu{} & {\bf 78.5}\sigSSS & {\bf 66.7}\sigSSS & {\bf 66.0}\sigSSS &  63.2\sigSSS &   63.2\sigSSS & {\bf 62.5}\sigSSS & {\bf 56.9}\sigSSS & {\bf 62.5}\sigSSS & {\bf 58.3}\sigSSS &  53.5\sigSS & {\bf 59.7}\sigSS  \\
& \emuNoLD{} & 76.4\sigSSS &   63.2\sigSSS &   59.7\sigSSS & {\bf 65.3}\sigSSS & {\bf 66.7}\sigSSS &   56.9\sigSSS &   55.6\sigSSS &   61.8\sigSSS &   56.9 &   54.2\sigSS &   58.3\sigSS \\
& \emuNoLDCL{} & 77.1\sigSSS &   62.5\sigSSS &   60.4\sigSSS &   61.8\sigSSS &   68.1\sigSSS &   58.3\sigSSS &   58.3\sigSSS & {\bf 62.5}\sigSSS &   56.9\sigSSS & {\bf 55.6}\sigSSS &   58.3\sigSS \\\cline{2-13}
& \emup{} & \underline{79.2}\sigSSS & \underline{68.1}\sigSSS &   65.3\sigSSS &  \underline{65.3}\sigSSS &   59.0\sigSSS &   61.8\sigSSS & \underline{59.7}\sigSSS &   61.1\sigSSS & \underline{59.0}\sigSS &  48.6 &   58.3\sigSS  \\\hline
\end{tabular}

\vspace{1em}
(b) \atis{}\\
    \centering
    \begin{tabular}{l|l|p{0.8cm}|p{0.8cm}p{0.8cm}p{0.8cm}p{0.8cm}p{0.8cm}|p{0.8cm}p{0.8cm}p{0.8cm}p{0.8cm}p{0.8cm}}\hline
                & & & \multicolumn{5}{c|}{{\sc En} $\to$ *} & \multicolumn{5}{c}{* $\to$ {\sc En}} \\\hline
                & Method & en-en & de & es & fr & ja & zh & de & es & fr & ja & zh \\\hline
\multirow{5}{*}{\rotatebox{90}{Baseline}} & MT + sent2vec                 &   90.5 &   87.3 &   89.7 &   87.7 &    2.4 &    7.1 &   84.9 &   84.9 &   86.1 &   80.6 &   81.7 \\
& LASER (original)              &   88.5 &   86.5 &   84.1 &   81.3 &   85.3 &   87.7 &   87.7 &   87.7 &   85.7 &  86.5 & {\bf 87.7} \\\cline{2-13}
& Contrastive loss              &   83.3 &   62.3 &   67.9 &   63.9 &   44.0 &   52.8 &   66.7 &   69.4 &   64.3 &   57.9 &   59.1  \\
& N-pair loss                   &   81.0 &   57.1 &   49.2 &   52.0 &   30.2 &   42.1 &   58.3 &   57.9 &   55.2 &   41.3 &   41.3 \\
& Softmax loss                  &   90.5 &   48.0 &   63.5 &   52.0 &   56.0 &   52.0 &   50.0 &   46.0 &   45.2 &   35.7 &   39.7 \\\hline
\multirow{4}{*}{\rotatebox{90}{Proposed}} & \emu{} & {\bf 98.8}\sigSSS & {\bf 98.4}\sigSSS & {\bf 98.4}\sigSSS &   93.7\sigSSS & {\bf 98.4}\sigSSS & {\bf 97.6}\sigSSS & {\bf 95.6}\sigSSS & {\bf 94.8}\sigSSS & {\bf 94.0}\sigSSS & {\bf 89.3} &   84.5 \\
& \emuNoLD & 97.6\sigSSS &   98.0\sigSSS &   96.8\sigSSS &   94.8\sigSSS &   95.6\sigSSS &   93.7\sigSSS &   92.5\sigSSS &   83.3 &   88.5 &   84.5 &   87.3 \\
& \emuNoLDCL & 98.4\sigSSS &   97.6\sigSSS &   98.0\sigSSS & {\bf 96.4}\sigSSS &   94.0\sigSSS &   95.6\sigSSS &   91.7\sigSS &   84.5 &   87.7 &   82.1 &   86.1  \\\cline{2-13}
& \emup{} &  \underline{99.2}\sigSSS & \underline{98.4}\sigSSS &   98.0\sigSSS &   95.2\sigSSS &   97.2\sigSSS &   96.0\sigSSS &   95.2\sigSSS & \underline{95.6}\sigSSS &  93.7\sigSSS &   77.8 &   83.7 \\\hline
    \end{tabular}
\vspace{1em}
\\ (c) \quora{}\\
    \begin{tabular}{l|l|p{0.8cm}|p{0.8cm}p{0.8cm}p{0.8cm}p{0.8cm}p{0.8cm}|p{0.8cm}p{0.8cm}p{0.8cm}p{0.8cm}p{0.8cm}}\hline
&                & & \multicolumn{5}{c|}{{\sc En} $\to$ *} & \multicolumn{5}{c}{* $\to$ {\sc En}} \\\hline
 &               Method & en-en & de & es & fr & ja & zh & de & es & fr & ja & zh \\\hline
\multirow{5}{*}{\rotatebox{90}{Baseline}} & MT + sent2vec                 &   77.6 &   74.0 &   75.8 &   73.5 &    1.8 &   72.6 &   70.4 &   70.4 &   69.5 &   70.4 &   71.3 \\
& LASER (original)              &   88.8 &   83.9 &   {\bf 86.5} &  {\bf 85.2} &   {\bf 86.5} &  85.2 &   79.8 &   79.8 &  {\bf 84.3} & {\bf 88.3} & {\bf 85.7} \\\cline{2-13}
& Contrastive loss              &   65.0 &   35.4 &   43.0 &   42.6 &   27.8 &   26.0 &   50.2 &   59.2 &   54.3 &   50.7 &   49.8 \\
& N-pair loss                   &   61.4 &   23.8 &   40.4 &   35.9 &   12.6 &   26.5 &   50.2 &   53.4 &   45.7 &   50.7 &   52.0 \\
& Softmax loss                  &   75.8 &   20.2 &   35.4 &   30.5 &   12.1 &   16.1 &   31.4 &   39.0 &   35.9 &   28.3 &   26.0 \\\hline
\multirow{4}{*}{\rotatebox{90}{Proposed}} & \emu{} & {\bf 89.7} &  {\bf 87.0}$^{*}$ & {\bf 86.5} &  84.3 & {\bf 86.5} &   86.1 & {\bf 81.2} & {\bf 87.0} &  82.5 &   83.9 &   84.8 \\
& \emuNoLD{} &  89.7 &   83.9 &   85.7 &   83.4 &   82.1 & {\bf 87.4} &   77.6 &   84.3 & 83.4 &  86.5 &   83.9 \\
& \emuNoLDCL{} & 88.3 &   75.3 &   80.3 &   75.8 &   70.9 &   78.5 &   72.6 &   82.1 &   75.3 &   81.6 &   80.7  \\\cline{2-13}
& \emup{} &  86.5 &   78.0 &   79.4 &   76.2 &   74.4 &   78.5 &   78.9 &   82.1 &   78.0 &   80.3 &   82.1 \\\hline
    \end{tabular}
\end{table*}

\begin{table*}[th]
    \centering
    \caption{Relative performance (Acc@1 on \trustyou{}) of \emuNoLD{} models trained on different training languages against the original LASER model for each language pair.}\label{table:results_multilingual}    
    \footnotesize
    \begin{tabular}{l|rrr|rrr|rrr}\hline
Training data   & en-en & en-de & en-fr & de-en & de-de & de-fr & fr-en & fr-de & fr-fr \\\hline
 En only      & +37.5\% & +40.0\% & +34.3\% & +27.0\% & +10.0\% &  +1.7\% & +12.3\% & +12.7\% & +11.1\% \\
 De only      & +26.2\% & +47.7\% & +10.0\% & +49.2\% & +10.0\% & +25.0\% &  +9.6\% &  +7.9\% &  +9.9\% \\
 Fr only      & +30.0\% & +33.8\% & +28.6\% & +17.5\% &  +8.7\% & +16.7\% & +31.5\% & +15.9\% & +17.3\% \\ \hline
 En + De      & +37.5\% & +58.5\% & +27.1\% & +50.8\% & +17.5\% & +23.3\% &  +9.6\% & +12.7\% & +14.8\% \\
 En + Fr      & +40.0\% & +60.0\% & {\bf +50.0\%} & +46.0\% & +12.5\% & +33.3\% & {\bf +35.6\%} & +25.4\% & +23.5\% \\ 
 De + Fr      & +28.7\% & +50.8\% & +37.1\% & +55.6\% & +12.5\% & +46.7\% & +31.5\% & +25.4\% & +17.3\% \\ \hline
 En + De + Fr & {\bf +41.2\%} & {\bf +63.1\%} & +47.1\% & {\bf +60.3\%} & {\bf +20.0\%} & {\bf +56.7\%} & +31.5\% & {\bf +34.9\%} & {\bf +25.9\%} \\ \hline
    \end{tabular}
    \normalsize
\end{table*}

\begin{table}[th]
    \caption{Relative performance of Acc@1 on \trustyou{} of \emuNoLD{} against the original LASER model for each language pair.}
    \label{table:results_langpairs}    
    \footnotesize
\begin{tabular}{c|P{0.8cm}|P{0.8cm}P{0.8cm}P{0.8cm}P{0.8cm}P{0.8cm}} \hline
*$\to$   & \multicolumn{1}{c}{en} & de & es & fr & zh & ja \\\hline
en       & \multicolumn{1}{c}{+37.5\%} & +40.0\% & +22.9\% & +34.3\% & +39.1\% & +26.1\% \\ \cline{3-7}
de       & +27.0\% & +10.0\% & +10.0\% & +1.7\% & +14.5\% & +20.9\% \\
es       & +34.8\% &  +0.0\% & +11.5\% &  +5.0\% & +21.3\% &  +8.0\% \\
fr       & +12.3\% & +12.7\% & +23.2\% & +11.1\% & +13.2\% &  +7.1\% \\
zh       & +21.9\% & +34.5\% & +11.4\% &  +9.6\% &  +9.1\% & +10.1\% \\
ja       & +18.3\% & +31.0\% & +23.4\% & +20.3\% & +32.3\% & +22.1\% \\\hline
    \end{tabular}
    \normalsize
\end{table}

\subsection{Dataset}
We used three datasets
for evaluation. Some statistics of these datasets are shown in Table~\ref{tab:dataset_stats}.

{\bf \trustyou{}} is a real-world private corpus of 820 questions collected via a multi-channel communication platform for hotel guests and hotel staff. Questions are always made by guests and have ground truth labels for 28 intent classes (e.g., check-in, pool.)
The utterances are professionally translated into 5 non-English languages (German (de), Spanish (es), French (fr), Japanese (ja), and Chinese (zh).)
We split the dataset into training and test sets so that the sentences used for fine-tuning do not appear in the test set.

{\bf \atis{}}~\cite{Hemphill:1990:ATIS} is a publicly available corpus for spoken dialog systems and is widely used for intent classification research. The dataset consists of more than 5k sentences and 22 intent labels are assigned to each sentence.
We excluded the ``flights'' class from the dataset since the class accounts for about 75\% of the dataset. We also ensured that each class has at least 5 sentences in each of train and test datasets. As a result, 13 classes remained in the dataset. %
Similar to previous studies~\cite{Conneau:2017:WordTranslationWOParallelData,Glavas:2019:HowToProperly}, we used Google Translate to generate corresponding translations in the same 5 non-English languages as \trustyou{}.

{\bf \quora{}}\footnote{\url{https://data.quora.com/First-Quora-Dataset-Release-Question-Pairs}} is a publicly available paraphrase detection dataset that contains over 400k questions with duplicate labels. Each row is a pair of questions with a duplicate label. 
Duplicate questions can be considered sentences that belong to the same intent. Therefore, we created a graph where each node is a question and an edge between two nodes denotes that these questions are considered duplicate. By doing this, we can consider each disjoint clique in the graph as a single intent class. Specifically, we filtered only complete subgraphs whose size (i.e., \# of nodes) is less than 30 to avoid having extremely large clusters that are too general. We chose the 50 largest clusters after the filtering. 
The original dataset contains only English sentences. We used Google Translate to translate into the same 5 languages in the same manner as \atis{}.

\subsection{Baselines}
\noindent
{\bf MT + sent2vec} We consider the two-stage approach that uses machine translation and monolingual sentence embeddings in a pipeline\footnote{The non-English sentences obtained through MT from English had to be translated back to English.}
%
We used Google Translate for translation and sent2vec~\cite{Pagliardini:2017:sent2vec} as a baseline method\footnote{We tested the official implementation of InferSent~\cite{Conneau:2017:InferSent}, finding that performance was unstable and often significantly lower than that of sent2vec. Thus, we decided to use sent2vec in the experiments.}.

\noindent
{\bf Softmax loss} Softmax loss is the most common loss function for classification, and thus a natural choice for fine-tuning the embeddings. We used the softmax loss function to train the semantic classifier and adjust the embeddings.

\noindent
{\bf Contrastive loss} Contrastive loss~\cite{Chopra:2005:SiameseOriginal} is a widely used pairwise loss function for metric learning. The loss function minimizes the squared distance between two embeddings if the labels are the same, and it maximizes the margin (we used $m=2.0$) between two samples otherwise. 
For contrastive loss, we use the Siamese (i.e., dual-encoder) architecture~\cite{Chopra:2005:SiameseOriginal} that takes two input sentences that will be fed into a shared encoder (i.e., multilingual encoder $E$) to obtain sentence embeddings.

\noindent
{\bf N-pair loss}
As another metric learning method, we used the N-pair sampling cosine loss~\cite{Yang:2019:AdditiveMargin}, which first samples one positive sample and $N - 1$ negative samples and then minimizes a cosine similarity-based loss function.

\subsection{Experimental Settings}
For each dataset, we used only English training data to fine-tune the models with \emu{} and the baseline methods. To train \emu's language discriminator, we used unlabeled training data in other non-English languages (i.e., de, es, fr, ja, zh.)

\noindent
{\bf Emu variants} To verify the effect of the language discriminator and the center loss, we also evaluated \emu{} without the language discriminator (\emuNoLD) and \emu{} without the language discriminator or the center loss (\emuNoLDCL) as a part of an ablation study. Finally, we evaluated \emup{}, which uses {\it parallel sentences} instead of randomly sampled sentences for cross-lingual adversarial training.

\noindent
{\bf Hyper-parameters} We used the official implementation of LASER\footnote{\url{https://github.com/facebookresearch/LASER}} and the pre-trained models including BPE. We implemented our proposed method and the baseline methods using PyTorch. 
We used an initial learning rate of $10^{-3}$ and optimized the model with Adam. We used a batch size of 16. For our proposed methods, we set $\alpha=50$ and $\lambda=10^{-4}$. All the models were trained for 3 epochs. The architecture of language discriminator $D$ has two 900-dimensional fully-connected layers with a dropout rate of 0.2. The hyper-parameters were $\gamma=10^{-4}$, $k=5$, $c=0.01$ respectively. The language discriminator was also optimized with Adam with an initial learning rate of $5.0 \times 10^{-4}$.

\noindent
{\bf Evaluation Metric} We used the leave-one-out evaluation method on the test data. For each sentence, we consider the other sentences in the test data as labeled sentences to find the nearest neighbor to predict the label. The idea is to exclude the direct translation of an input sentence in the target language to make the nearest neighbor search more challenging and to simulate the real-world setting where parallel sentences are missing. We used Acc@1 (the ratio of test sentences that are correctly categorized into the intent classes) as our evaluation metric.

\subsection{Results and Discussion}
Table \ref{table:results_ty} shows the experimental results on these three datasets. 
In Table \ref{table:results_ty} (a), \emu{} achieved the best performance for all the 11 tasks (en-fr, en-ja, and ja-en by \emuNoLD{} and en-ja by \emuNoLDCL{}.) \emu{} outperformed the baseline methods including the original LASER model. 
In Table \ref{table:results_ty} (b), \emu{} achieved the best performance for 10 tasks (en-fr by \emuNoLDCL.) The original LASER model showed the best performance for zh-en and all of the \emu{} methods degraded the performance for the task.
In Table \ref{table:results_ty} (c), \emu{} achieved the best performance for 7 tasks (en-zh by \emuNoLD{}), whereas the original LASER model achieved the best performance for the rest of the tasks.
From the results, \emu{} consistently outperformed the baseline methods, including the original LASER model. 
At the same time, \emu{} failed to improve the performance of the five tasks, namely zh-en on \atis{} (Table~\ref{table:results_ty} (b)) and en-fr, fr-en, ja-en, ja-zh on \quora{} (Table~\ref{table:results_ty} (c)).
We would like to emphasize that the \emu{} models were trained using labeled data only in English. The \emu{} also used unlabeled data in non-English languages. Therefore, it is noteworthy that our framework successfully specializes multilingual sentence emebeddings for multiple language pairs, which involve English, using only English labeled data. The results support that \emu{} is effective in semantically specializing multilingual sentence embeddings.
%

For all the tasks, we observe that the baseline fine-tuning methods (i.e., contrastive loss, N-pair loss, softmax loss) do not improve the performance but instead decrease the accuracy values compared to the original LASER performance. The results indicate that fine-tuning multilingual sentence embeddings is sensitive to the choice of loss functions, and $L_2$-constrained softmax loss is the best choice among the loss functions. 

\begin{table}[t!]
    \centering
    \caption{Ablation study of \emu. Each value denotes the average percentage point (pp) drop after removing the component. Negative values denote {\it improvements} after removing the component. \sigSS and \sigSSS denote $p$-values $<0.05$ and $<0.01$ (Wilcoxon signed ranked test) respectively.}\label{table:ablation_analysis}   
\footnotesize    
    \begin{tabular}{c||l|l|l}\hline
        Component & \trustyou{} & \atis{} & \quora{}  \\\hline
        Language Discriminator & $1.45$ & $2.81$\sigSS & $1.05$ \\
        Center loss & $-0.44$ & $0.04$ & $6.00$\sigSSS \\\hline
    \end{tabular}
\normalsize
\end{table}

\hl{The original LASER model consistently performs better on all datasets for the en-en task compared to the other tasks. This is partially due to the higher quality sentence embeddings in English. More specifically, the LASER model was trained on MT tasks, translating text from 93 languages to either English or Spanish as target languages with English having the most training data in the dataset~\cite{Artetxe:2018:LASER}.}

\hl{MT+sent2vec shows significantly low Acc@1 values for the en-ja and en-zh tasks on \atis{}, and for the en-ja task on \quora{}. Investigating this trend, we observed that back-translation of sentences that were translated from en into ja/zh results in the following types of degradation: (1) missing words, especially interrogative pronouns (e.g., what, when, which etc.) and verbs, (2) significant changes in the word order. As discussed above, sent2vec embeddings are also susceptible to this type of perturbation. From the evaluation perspective, we can consider the en-en performance of MT+sent2vec as an upper bound on its performance for all en-* and *-en tasks, assuming that MT exactly translates back to the original sentence in English. Nevertheless, \emu{} consistently outperforms MT+sent2vec.}

\noindent {\bf Ablation study} We conducted an ablation study to quantitatively evaluate the contribution of each component of \emu{}, namely, the language discriminator and the center loss. First, we compared \emuNoLD{} with \emu{} to verify the effect of the language discriminator, and then compared \emuNoLD{} and \emuNoLDCL{} to determine the effect of the center loss. 

Table \ref{table:ablation_analysis} shows the average percentage point drop (i.e., the degree of contributions) of each component. The language discriminator had a significant contribution of 2.81 points on \atis{}. The contributions were 1.45 points and 1.05 points on \trustyou{} and \quora{} respectively. 
Similarly, the center loss had a significant impact on \quora{}, whereas it had almost no effect on \atis{} and had a negative impact on \trustyou{}.
%

\begin{figure*}[th]
      \centering
            \subfigure[]		{ \includegraphics[width=0.23\textwidth]{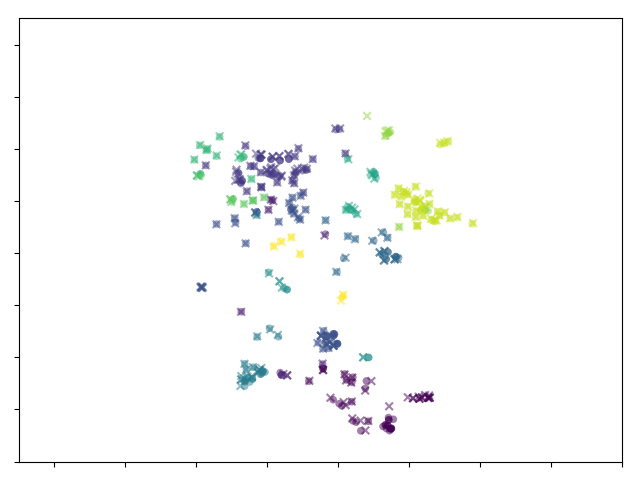} \label{fig:pca_original}}
            \subfigure[] 		{ \includegraphics[width=0.23\textwidth]{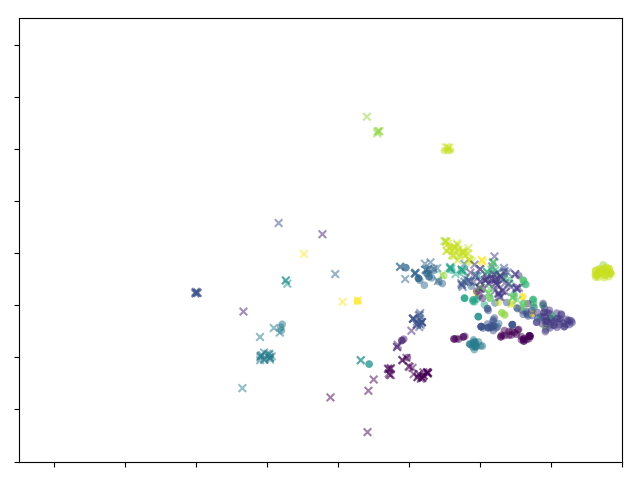}		\label{fig:pca_softmax}}
            \subfigure[]		{ \includegraphics[width=0.23\textwidth]{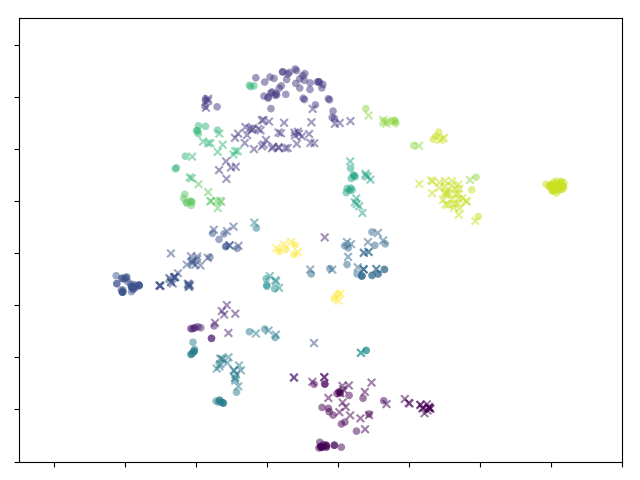}\label{fig:pca_proposed_wo_ld}}
            \subfigure[] 		{ \includegraphics[width=0.23\textwidth]{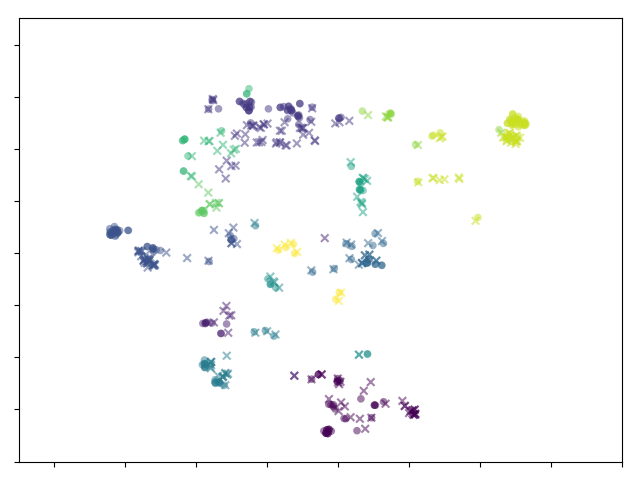} \label{fig:pca_proposed}}
        \caption{Visualizations of the sentence embeddings of English ($\circ$) and German ($\times$) test data of the \atis{} dataset. We used $t$-SNE to convert the sentence embeddings into the 2d space. Each point is a sentence and the color denotes the intent class. The plots are: (a) the original LASER embeddings, (b) softmax loss, (c) \emuNoLD, (d) \emu.}
        \label{fig:pca_analysis_methods}
\end{figure*}

\noindent
{\bf Sentence embedding visualization}
We conducted a qualitative analysis to observe how our framework with the language discriminator specialized multilingual sentence embeddings and enhanced the multilinguality. We filtered English and German sentences from the test data of the \atis{} dataset and visualized sentence embeddings of (a) the original LASER model, (b) the softmax loss, (c) \emuNoLD, and (d) \emu{} into the same 2D space using $t$-SNE.

Figure \ref{fig:pca_analysis_methods} shows visualizations of these methods. Figure \ref{fig:pca_original} shows that the original LASER sentence embeddings have multilinguality, as the sentences in the same intent in English and German were embedded close to each other. Figure \ref{fig:pca_softmax} shows that fine-tuning the model with the softmax loss function broke not only the intent clusters but also spoiled the multilinguality. In Figure \ref{fig:pca_proposed_wo_ld}, \emuNoLD{} successfully specialized the sentence embeddings, whereas multilinguality was degraded as the sentence embeddings of the same intent classes were separated compared to the original LASER model. Finally, \emu{} (with the language discriminator) moved sentence embeddings of the same intent in English and German close to each other, as shown in Figure \ref{fig:pca_proposed}.

From the results, we observe that incorporating the language discriminator enriches the multilinguality in the embedding space.

\noindent {\bf Do we need parallel sentences for Emu?}
We compared \emu{} to \emup{}, which uses parallel sentences instead of randomly sampled sentences, to verify whether using parallel sentences makes multilingual adversarial learning more effective. The results are shown in Tables \ref{table:results_ty} (a)-(c). 
Compared to \emu{}, \emup{} showed lower Acc@1 values on the three datasets. The decreases were -0.5 points, -1.2 points, and -5.9 points on \trustyou{}, \atis{}, and \quora{} respectively.  The differences are not statistically significant except for \quora{}. 
The results show that the language discriminator of \emu{} does not need any cost-expensive parallel corpus but can improve performance using {\it unlabeled and non-parallel} sentences in other languages.

\noindent
{\bf What language(s) should we use for training?}
We also investigated how the performance changes by fine-tuning with training data in multiple languages other than English. To understand the insights more closely, we turned off the language discriminator in this analysis to ensure that \emu{} uses data only in specified languages. 
We summarize the relative performance of \emuNoLD{} against the original LASER model on the \trustyou{} dataset. As discussed above, the accuracy values of tasks that involve English in at least one side (i.e., source language, target language, or both) show larger improvements than the other pairs that only involve non-English languages. This is likely because sentence embeddings of those languages were not appropriately fine-tuned compared to those of English because training data in those languages were not used. 

Therefore, we hypothesized that using training data in the same language for a target and/or source language would be the best choice. To test the hypothesis, we chose English, German, and French as source/target languages and conducted additional experiments on the \trustyou{} dataset. The experimental settings, including the hyper-parameters, followed the main experiments, with only the training data used for fine-tuning being different.

Table~\ref{table:results_multilingual} shows the results. When only using training data in a single language (i.e., En only, De only, Fr only), the target language was the best training data for monolingual intent classification tasks because this method achieved the best performance in the en-en, de-de, and fr-fr tasks respectively. Similarly, using the source and target languages as training data was the best configuration for methods that trained in two languages. That is, En+De achieved the best performance for the en-de and de-en tasks. En+Fr (De+Fr) also achieved the best performance for the en-fr (de-fr) and fr-en (fr-de.)
Finally, the method that used training data in the three languages (En+De+Fr) showed the best accuracy values for 7 out of 9 tasks. The degradation in those two tasks occurred when En+De+Fr incorporated a language that was neither the source nor target languages (i.e., en-fr and fr-en.)

From the results, we conclude that we should focus on creating training data in a target or source language to obtain the best performance with \emu{} and use our budget effectively.

\section{Related Work}
Multilingual embedding techniques~\cite{Ruder:2019:CLSurvey} have been well studied, and most of the prior work has focused on word embeddings.
%
However, relatively fewer techniques have been developed for multilingual sentence embeddings. This is because such techniques~\cite{Hermann:2014:BiCVM,Artetxe:2018:LASER} require parallel sentences for training multilingual sentence embeddings and some use both sentence-level and word-level alignment information~\cite{Luong:2015:BiSkip}.
\hl{Ruckle et al.~\cite{Ruckle:2018:Concatenated}
developed an unsupervised sentence embedding method based on concatenating and aggregating cross-lingual word embeddings. They also confirm that the method performs well on cross-lingual as well as monolingual settings.}
\hl{Schwenk and Douze~\cite{Schwenk:2017:LearningJointMultilingual}
used machine translation tasks to learn multilingual sentence representations.}\hl{ This idea has been further expanded in LASER~\cite{Artetxe:2018:LASER,Artetxe:2018:MarginParallelCorpusMining}, a recently developed system which trains a language-agnostic sentence embeddings model with a large number of translation tasks on a large-scale parallel corpora.}

Similar to the center loss used in this paper, two techniques have incorporated cluster-level information~\cite{Huang:2018:EMNLP,Doval:2018:MeetingInTheMiddle} to enhance the compactness of word clusters to improve the quality of multilingual word embedding models. None of them have directly used the centroid of each class to calculate loss values for training.

Adversarial learning~\cite{Goodfellow:2014:GAN} is a common technique that has been used for many NLP tasks, including (monolingual) sentence embeddings~\cite{Patro:2018:PairwiseDiscriminator} and multilingual word embeddings~\cite{Conneau:2017:WordTranslationWOParallelData,Chen:2018:UnsupervisedMWE}. \cite{Chen:2018:ADAN} developed a technique that uses a language discriminator to train a cross-lingual sentiment classifier. Our framework is similar in the use of a language discriminator, but our novelty is that it uses a language discriminator for learning multilingual sentence embeddings instead of cross-lingual transfer.
\hl{Joty et al.~\cite{Joty:2017:CrossLanguage} used a language discriminator to train a model for cross-lingual question similarity calculation. Their setting differs from ours as their method requires parallel sentences in different languages and pair-wise similarity labels instead of class labels.}
 
There is a line of work in post-processing word embedding models called {\it word embedding specialization}~\cite{Faruqui:2015:RetrofittingWE,Kiela:2015:SpecializingWE,Mrksic:2017:SemanticSpecialization}.
Prior work specialized word embeddings with different external resources such as semantic information~\cite{Faruqui:2015:RetrofittingWE}.
The common approaches are (1) a post-hoc learning~\cite{Faruqui:2015:RetrofittingWE} that uses additional loss function to tune pre-trained embeddings, (2) learning an additional model~\cite{Glavas:2018:ExplicitRetrofitting,Vulic:2018:PostSpecialisation}, and (3) the fine-tuning approach~\cite{Abdalla:2019:Enriching}, which is similar to our fine-tuning approach.
However, to the best of our knowledge, we are the first to approach semantic specialization of multilingual sentence embeddings.

\section{Conclusion}
We have presented \emu{}, a semantic specialization framework for multilingual sentence embeddings. \emu{} incorporates multilingual adversarial training on top of fine-tuning to enhance multilinguality without using parallel sentences.

Our experimental results show that \emu{} outperformed the baseline methods including state-of-the-art multilingual sentence emebeddings, LASER, and monolingual sentence embeddings after machine translation with respect to multiple language pairs. 
The results also show that \emu{} can successfully train a model using only monolingual labeled data and unlabeled data in other languages.

\section{Acknowledgments}
\hl{We thank Sorami Hisamoto for sharing his literature survey on cross-lingual embedding techniques and Tom Mitchell for helpful comments as well as the anonymous reviewers for their constructive feedback.}

\balance
\bibliography{ref}

\begin{thebibliography}{}

\bibitem[\protect\citeauthoryear{Abdalla, Sahlgren, and
  Hirst}{2019}]{Abdalla:2019:Enriching}
Abdalla, M.; Sahlgren, M.; and Hirst, G.
\newblock 2019.
\newblock Enriching word embeddings with a regressor instead of labeled
  corpora.
\newblock In {\em Proc. AAAI '19},  6188--6195.

\bibitem[\protect\citeauthoryear{Arjovsky, Chintala, and
  Bottou}{2017}]{Arjovsky:2017:WGAN}
Arjovsky, M.; Chintala, S.; and Bottou, L.
\newblock 2017.
\newblock {W}asserstein generative adversarial networks.
\newblock In {\em Proc. ICML '17}, volume~70,  214--223.

\bibitem[\protect\citeauthoryear{Artetxe and
  Schwenk}{2019a}]{Artetxe:2018:MarginParallelCorpusMining}
Artetxe, M., and Schwenk, H.
\newblock 2019a.
\newblock Margin-based parallel corpus mining with multilingual sentence
  embeddings.
\newblock In {\em Proc. ACL '19},  3197--3203.

\bibitem[\protect\citeauthoryear{Artetxe and
  Schwenk}{2019b}]{Artetxe:2018:LASER}
Artetxe, M., and Schwenk, H.
\newblock 2019b.
\newblock Massively multilingual sentence embeddings for zero-shot
  cross-lingual transfer and beyond.
\newblock {\em Transactions of the Association for Computational Linguistics}
  7:597--610.

\bibitem[\protect\citeauthoryear{Chen and
  Cardie}{2018}]{Chen:2018:UnsupervisedMWE}
Chen, X., and Cardie, C.
\newblock 2018.
\newblock Unsupervised multilingual word embeddings.
\newblock In {\em Proc. EMNLP '18},  261--270.

\bibitem[\protect\citeauthoryear{Chen \bgroup et al\mbox.\egroup
  }{2018}]{Chen:2018:ADAN}
Chen, X.; Sun, Y.; Athiwaratkun, B.; Cardie, C.; and Weinberger, K.
\newblock 2018.
\newblock Adversarial deep averaging networks for cross-lingual sentiment
  classification.
\newblock {\em Transactions of the Association for Computational Linguistics}
  6:557--570.

\bibitem[\protect\citeauthoryear{Chidambaram \bgroup et al\mbox.\egroup
  }{2019}]{Chidambaram:2018:CrossLingUSE}
Chidambaram, M.; Yang, Y.; Cer, D.; Yuan, S.; Sung, Y.; Strope, B.; and
  Kurzweil, R.
\newblock 2019.
\newblock Learning cross-lingual sentence representations via a multi-task
  dual-encoder model.
\newblock In {\em Proc. {RepL4NLP} '19},  250--259.

\bibitem[\protect\citeauthoryear{Chopra \bgroup et al\mbox.\egroup
  }{2005}]{Chopra:2005:SiameseOriginal}
Chopra, S.; Hadsell, R.; LeCun, Y.; et~al.
\newblock 2005.
\newblock Learning a similarity metric discriminatively, with application to
  face verification.
\newblock In {\em Proc. CVPR '05},  539--546.

\bibitem[\protect\citeauthoryear{Conneau \bgroup et al\mbox.\egroup
  }{2017}]{Conneau:2017:InferSent}
Conneau, A.; Kiela, D.; Schwenk, H.; Barrault, L.; and Bordes, A.
\newblock 2017.
\newblock Supervised learning of universal sentence representations from
  natural language inference data.
\newblock In {\em Proc. EMNLP '17},  670--680.

\bibitem[\protect\citeauthoryear{Conneau \bgroup et al\mbox.\egroup
  }{2018a}]{Conneau:2018:WhatYouCanCram}
Conneau, A.; Kruszewski, G.; Lample, G.; Barrault, L.; and Baroni, M.
\newblock 2018a.
\newblock What you can cram into a single {\$}{\&}!{\#}* vector: Probing
  sentence embeddings for linguistic properties.
\newblock In {\em Proc. ACL '18},  2126--2136.

\bibitem[\protect\citeauthoryear{Conneau \bgroup et al\mbox.\egroup
  }{2018b}]{Conneau:2017:WordTranslationWOParallelData}
Conneau, A.; Lample, G.; Ranzato, M.; Denoyer, L.; and J{\'e}gou, H.
\newblock 2018b.
\newblock Word translation without parallel data.
\newblock In {\em Proc. ICLR '18}.

\bibitem[\protect\citeauthoryear{Devlin \bgroup et al\mbox.\egroup
  }{2019}]{Devlin:2018:BERT}
Devlin, J.; Chang, M.-W.; Lee, K.; and Toutanova, K.
\newblock 2019.
\newblock {BERT}: Pre-training of deep bidirectional transformers for language
  understanding.
\newblock In {\em Proc. NAACL-HLT '19},  4171--4186.

\bibitem[\protect\citeauthoryear{Doval \bgroup et al\mbox.\egroup
  }{2018}]{Doval:2018:MeetingInTheMiddle}
Doval, Y.; Camacho-Collados, J.; Espinosa~Anke, L.; and Schockaert, S.
\newblock 2018.
\newblock Improving cross-lingual word embeddings by meeting in the middle.
\newblock In {\em Proc. EMNLP '18},  294--304.

\bibitem[\protect\citeauthoryear{Faruqui \bgroup et al\mbox.\egroup
  }{2015}]{Faruqui:2015:RetrofittingWE}
Faruqui, M.; Dodge, J.; Jauhar, S.~K.; Dyer, C.; Hovy, E.; and Smith, N.~A.
\newblock 2015.
\newblock Retrofitting word vectors to semantic lexicons.
\newblock In {\em Proc. NAACL-HLT '15},  1606--1615.

\bibitem[\protect\citeauthoryear{Glava{\v{s}} and
  Vuli{\'c}}{2018}]{Glavas:2018:ExplicitRetrofitting}
Glava{\v{s}}, G., and Vuli{\'c}, I.
\newblock 2018.
\newblock Explicit retrofitting of distributional word vectors.
\newblock In {\em Proc. ACL '18},  34--45.

\bibitem[\protect\citeauthoryear{Glavas \bgroup et al\mbox.\egroup
  }{2019}]{Glavas:2019:HowToProperly}
Glavas, G.; Litschko, R.; Ruder, S.; and Vulic, I.
\newblock 2019.
\newblock How to (properly) evaluate cross-lingual word embeddings: On strong
  baselines, comparative analyses, and some misconceptions.
\newblock In {\em Proc. ACL '19}.

\bibitem[\protect\citeauthoryear{Goodfellow \bgroup et al\mbox.\egroup
  }{2014}]{Goodfellow:2014:GAN}
Goodfellow, I.; Pouget-Abadie, J.; Mirza, M.; Xu, B.; Warde-Farley, D.; Ozair,
  S.; Courville, A.; and Bengio, Y.
\newblock 2014.
\newblock Generative adversarial nets.
\newblock In {\em Proc. NIPS '14},  2672--2680.

\bibitem[\protect\citeauthoryear{Hemphill, Godfrey, and
  Doddington}{1990}]{Hemphill:1990:ATIS}
Hemphill, C.~T.; Godfrey, J.~J.; and Doddington, G.~R.
\newblock 1990.
\newblock The {ATIS} spoken language systems pilot corpus.
\newblock In {\em Proc. the Workshop on Speech and Natural Language}, HLT '90,
  96--101.

\bibitem[\protect\citeauthoryear{Hermann and
  Blunsom}{2014}]{Hermann:2014:BiCVM}
Hermann, K.~M., and Blunsom, P.
\newblock 2014.
\newblock Multilingual models for compositional distributed semantics.
\newblock In {\em Proc. ACL '14},  58--68.

\bibitem[\protect\citeauthoryear{Howard and Ruder}{2018}]{Jeremy:2018:ULMFiT}
Howard, J., and Ruder, S.
\newblock 2018.
\newblock Universal language model fine-tuning for text classification.
\newblock In {\em Proc. ACL '18},  328--339.

\bibitem[\protect\citeauthoryear{Huang \bgroup et al\mbox.\egroup
  }{2018}]{Huang:2018:EMNLP}
Huang, L.; Cho, K.; Zhang, B.; Ji, H.; and Knight, K.
\newblock 2018.
\newblock Multi-lingual common semantic space construction via
  cluster-consistent word embedding.
\newblock In {\em Prc. EMNLP '18}.

\bibitem[\protect\citeauthoryear{Joty \bgroup et al\mbox.\egroup
  }{2017}]{Joty:2017:CrossLanguage}
Joty, S.; Nakov, P.; M{\`a}rquez, L.; and Jaradat, I.
\newblock 2017.
\newblock Cross-language learning with adversarial neural networks.
\newblock In {\em Proc. {C}o{NLL} '17},  226--237.

\bibitem[\protect\citeauthoryear{Kiela, Hill, and
  Clark}{2015}]{Kiela:2015:SpecializingWE}
Kiela, D.; Hill, F.; and Clark, S.
\newblock 2015.
\newblock Specializing word embeddings for similarity or relatedness.
\newblock In {\em Proc. EMNLP '15},  2044--2048.

\bibitem[\protect\citeauthoryear{Luong, Pham, and
  Manning}{2015}]{Luong:2015:BiSkip}
Luong, T.; Pham, H.; and Manning, C.~D.
\newblock 2015.
\newblock Bilingual word representations with monolingual quality in mind.
\newblock In {\em Proc. RepL4NLP '15},  151--159.

\bibitem[\protect\citeauthoryear{Mrk{\v{s}}i{\'c} \bgroup et al\mbox.\egroup
  }{2017}]{Mrksic:2017:SemanticSpecialization}
Mrk{\v{s}}i{\'c}, N.; Vuli{\'c}, I.; {\'O}~S{\'e}aghdha, D.; Leviant, I.;
  Reichart, R.; Ga{\v{s}}i{\'c}, M.; Korhonen, A.; and Young, S.
\newblock 2017.
\newblock Semantic specialization of distributional word vector spaces using
  monolingual and cross-lingual constraints.
\newblock {\em Transactions of the Association for Computational Linguistics}
  5:309--324.

\bibitem[\protect\citeauthoryear{Pagliardini, Gupta, and
  Jaggi}{2018}]{Pagliardini:2017:sent2vec}
Pagliardini, M.; Gupta, P.; and Jaggi, M.
\newblock 2018.
\newblock Unsupervised learning of sentence embeddings using compositional
  n-gram features.
\newblock In {\em NAACL-HLT '18}.

\bibitem[\protect\citeauthoryear{Pa{\c{s}}ca}{2003}]{Pacsca:2003:OpenQA}
Pa{\c{s}}ca, M.
\newblock 2003.
\newblock {\em Open-domain question answering from large text collections}.
\newblock MIT Press.

\bibitem[\protect\citeauthoryear{Patro \bgroup et al\mbox.\egroup
  }{2018}]{Patro:2018:PairwiseDiscriminator}
Patro, B.~N.; Kurmi, V.~K.; Kumar, S.; and Namboodiri, V.~P.
\newblock 2018.
\newblock Learning semantic sentence embeddings using pair-wise discriminator.
\newblock In {\em Proc. COLING '18}.

\bibitem[\protect\citeauthoryear{Peters \bgroup et al\mbox.\egroup
  }{2018}]{Peters:2018:ELMo}
Peters, M.; Neumann, M.; Iyyer, M.; Gardner, M.; Clark, C.; Lee, K.; and
  Zettlemoyer, L.
\newblock 2018.
\newblock Deep contextualized word representations.
\newblock In {\em Proc. NAACL-HLT '18},  2227--2237.

\bibitem[\protect\citeauthoryear{Ranjan, Castillo, and
  Chellappa}{2017}]{Ranjan:2017:L2constrainedSL}
Ranjan, R.; Castillo, C.~D.; and Chellappa, R.
\newblock 2017.
\newblock L2-constrained softmax loss for discriminative face verification.
\newblock {\em arXiv prepring arXiv:1703.09507} abs/1703.09507.

\bibitem[\protect\citeauthoryear{Reimers and
  Gurevych}{2019}]{Reimers:2019:SentenceBERT}
Reimers, N., and Gurevych, I.
\newblock 2019.
\newblock Sentence-{BERT}: Sentence embeddings using {S}iamese {BERT}-networks.
\newblock In {\em Proc. EMNLP-IJCNLP '19},  3973--3983.

\bibitem[\protect\citeauthoryear{R{\"u}ckl{\'e} \bgroup et al\mbox.\egroup
  }{2018}]{Ruckle:2018:Concatenated}
R{\"u}ckl{\'e}, A.; Eger, S.; Peyrard, M.; and Gurevych, I.
\newblock 2018.
\newblock Concatenated power mean word embeddings as universal cross-lingual
  sentence representations.
\newblock {\em arXiv preprint arXiv:1803.01400}.

\bibitem[\protect\citeauthoryear{Ruder \bgroup et al\mbox.\egroup
  }{2019}]{Ruder:2019:CLSurvey}
Ruder, S.; Vuli{\'c}, I.; S{\o}gaard, A.; and Faruqui, M.
\newblock 2019.
\newblock {\em Cross-Lingual Word Embeddings}.
\newblock Morgan \& Claypool Publishers.

\bibitem[\protect\citeauthoryear{Schwenk and
  Douze}{2017}]{Schwenk:2017:LearningJointMultilingual}
Schwenk, H., and Douze, M.
\newblock 2017.
\newblock Learning joint multilingual sentence representations with neural
  machine translation.
\newblock In {\em Proc. {RepL4NLP} '17},  157--167.

\bibitem[\protect\citeauthoryear{Sennrich, Haddow, and
  Birch}{2016}]{Sennrich:2016:BPE}
Sennrich, R.; Haddow, B.; and Birch, A.
\newblock 2016.
\newblock Neural machine translation of rare words with subword units.
\newblock In {\em Proc. ACL '16},  1715--1725.

\bibitem[\protect\citeauthoryear{Vuli{\'c} \bgroup et al\mbox.\egroup
  }{2018}]{Vulic:2018:PostSpecialisation}
Vuli{\'c}, I.; Glava{\v{s}}, G.; Mrk{\v{s}}i{\'c}, N.; and Korhonen, A.
\newblock 2018.
\newblock Post-specialisation: {R}etrofitting vectors of words unseen in
  lexical resources.
\newblock In {\em Proc. NAACL-HLT '19}.

\bibitem[\protect\citeauthoryear{Wang \bgroup et al\mbox.\egroup
  }{2017}]{Wang:2017:LearningToHash}
Wang, J.; Zhang, T.; Sebe, N.; Shen, H.~T.; et~al.
\newblock 2017.
\newblock A survey on learning to hash.
\newblock {\em IEEE Transactions on On Pattern Analysis and Machine
  Intelligence} 40(4):769--790.

\bibitem[\protect\citeauthoryear{Wen \bgroup et al\mbox.\egroup
  }{2016}]{Wen:2016:CenterLoss}
Wen, Y.; Zhang, K.; Li, Z.; and Qiao, Y.
\newblock 2016.
\newblock A discriminative feature learning approach for deep face recognition.
\newblock In {\em Proc. ECCV '16},  499--515.

\bibitem[\protect\citeauthoryear{Yang \bgroup et al\mbox.\egroup
  }{2019}]{Yang:2019:AdditiveMargin}
Yang, Y.; Abrego, G.~H.; Yuan, S.; Guo, M.; Shen, Q.; Cer, D.; Sung, Y.-h.;
  Strope, B.; and Kurzweil, R.
\newblock 2019.
\newblock Improving multilingual sentence embedding using bi-directional dual
  encoder with additive margin softmax.
\newblock In {\em Proc. IJCAI '19},  5370--5378.

\bibitem[\protect\citeauthoryear{Zhu, Li, and de
  Melo}{2018}]{Zhu:2018:ExploringSemanticProp}
Zhu, X.; Li, T.; and de~Melo, G.
\newblock 2018.
\newblock Exploring semantic properties of sentence embeddings.
\newblock In {\em Proc. ACL '18},  632--637.

\end{thebibliography}
\bibliographystyle{aaai}

\end{document}